\def\year{2022}\relax
\documentclass[letterpaper]{article} 
\usepackage{aaai22}  
\usepackage{times}  
\usepackage{helvet}  
\usepackage{courier}  
\usepackage[hyphens]{url}  
\usepackage{graphicx} 
\usepackage{natbib}  
\usepackage{caption} 
\DeclareCaptionStyle{ruled}{labelfont=normalfont,labelsep=colon,strut=off} 
\usepackage{algorithm}
\usepackage{algorithmic}

\usepackage{epsfig}
\usepackage{amsmath}
\usepackage{amssymb}

\usepackage{microtype}
\usepackage{booktabs} 
\usepackage{multirow}
\usepackage{multicol}
\usepackage{color, colortbl}
\usepackage{adjustbox}
\usepackage{xspace}

\usepackage{array}
\newcolumntype{x}[1]{>{\centering\let\newline\\\arraybackslash\hspace{0pt}}p{#1}}

\usepackage{newfloat}
\usepackage{listings}
\floatstyle{ruled}
\newfloat{listing}{tb}{lst}{}
\floatname{listing}{Listing}
\title{Meta Faster R-CNN: Towards Accurate Few-Shot Object Detection with Attentive Feature Alignment}
\author{
    Guangxing Han, Shiyuan Huang, Jiawei Ma, Yicheng He, Shih-Fu Chang
}
\affiliations{
    Columbia University\\
    \{gh2561, sh3813, jiawei.m, yh3330, sc250\}@columbia.edu
}

\usepackage{bibentry}

\begin{document}

\maketitle

\begin{abstract}
Few-shot object detection (FSOD) aims to detect objects using only a few examples. How to adapt state-of-the-art object detectors to the few-shot domain remains challenging. Object proposal is a key ingredient in modern object detectors. However, the quality of proposals generated for few-shot classes using existing methods is far worse than that of many-shot classes, e.g., missing boxes for few-shot classes due to misclassification or inaccurate spatial locations with respect to true objects. To address the noisy proposal problem, we propose a novel meta-learning based FSOD model by jointly optimizing the few-shot proposal generation and fine-grained few-shot proposal classification. To improve proposal generation for few-shot classes, we propose to learn a lightweight metric-learning based prototype matching network, instead of the conventional simple linear object/nonobject classifier, e.g., used in RPN. Our non-linear classifier with the feature fusion network could improve the discriminative prototype matching and the proposal recall for few-shot classes. To improve the fine-grained few-shot proposal classification, we propose a novel attentive feature alignment method to address the spatial misalignment between the noisy proposals and few-shot classes, thus improving the performance of few-shot object detection. Meanwhile we learn a separate Faster R-CNN detection head for many-shot base classes and show strong performance of maintaining base-classes knowledge. Our model achieves state-of-the-art performance on multiple FSOD benchmarks over most of the shots and metrics.
\end{abstract}

\section{Introduction}

Object detection is one of the most fundamental and challenging tasks in computer vision. Recent years have witnessed great progress in this field using deep learning techniques \cite{ren2015faster,redmon2016you,liu2016ssd,he2017mask,tian2019fcos}. However, deep learning based object detection methods need a sufficient amount of human annotations for model training, which are expensive to collect and unavailable for rare categories. Given scarce training data, these models suffer from the risk of overfitting and poor generalization ability \cite{kang2019few}. 

\begin{figure}[t]
\begin{center}
\includegraphics[scale=0.25]{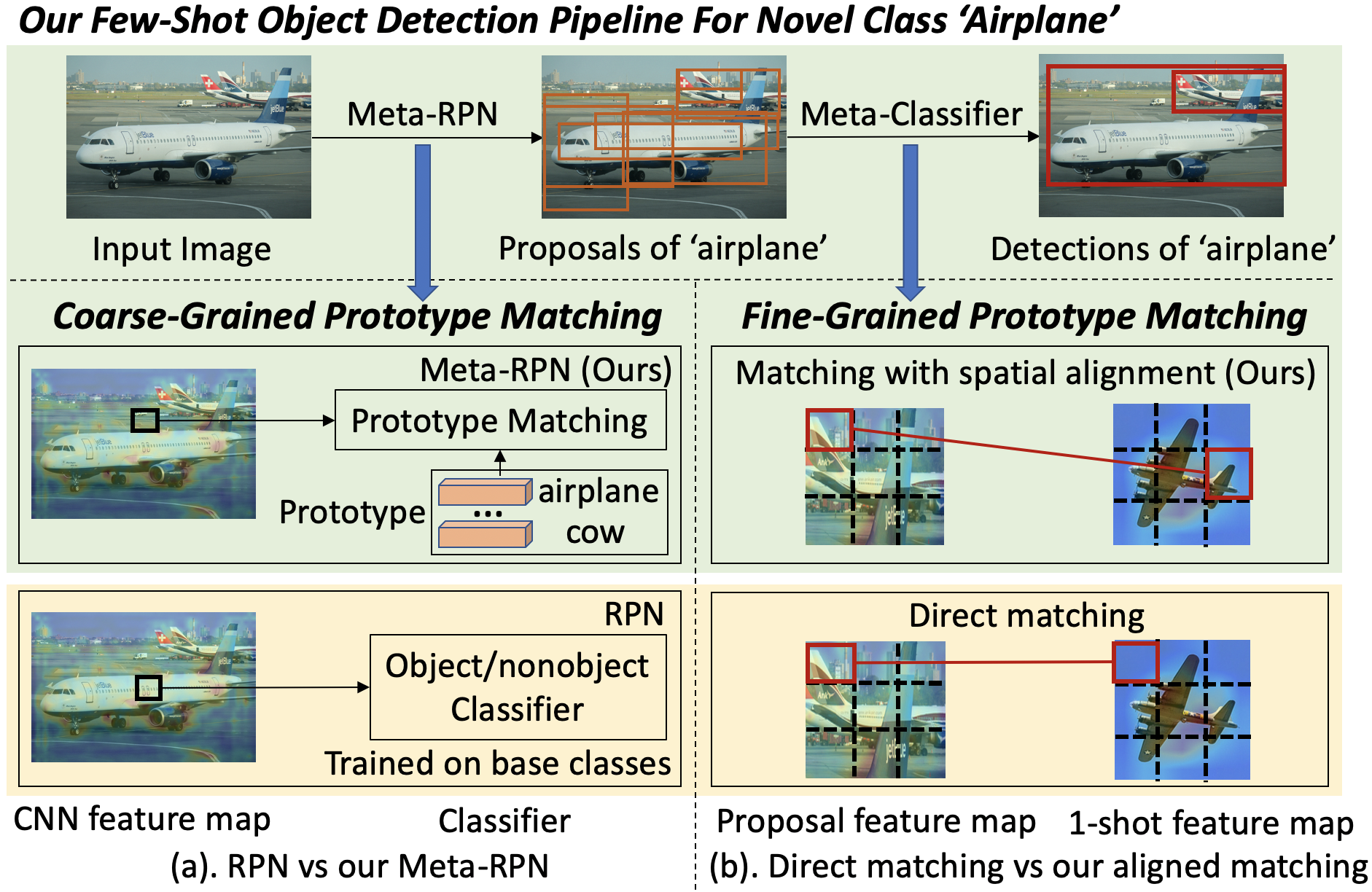}
\end{center}
\caption{Overview of the proposed model for FSOD. 
(a) We compare our coarse-grained prototype matching based Meta-RPN with the original RPN. Our method can generate category-specific proposals in an efficient and effective way. 
(b) Through feature alignment for fine-grained prototype matching, our method can address the spatial misalignment issues between proposals and few-shot classes, leading to better matching and FSOD performance. } 
\label{introduction}
\end{figure}

\begin{figure*}[t]
\begin{center}
\includegraphics[scale=0.35]{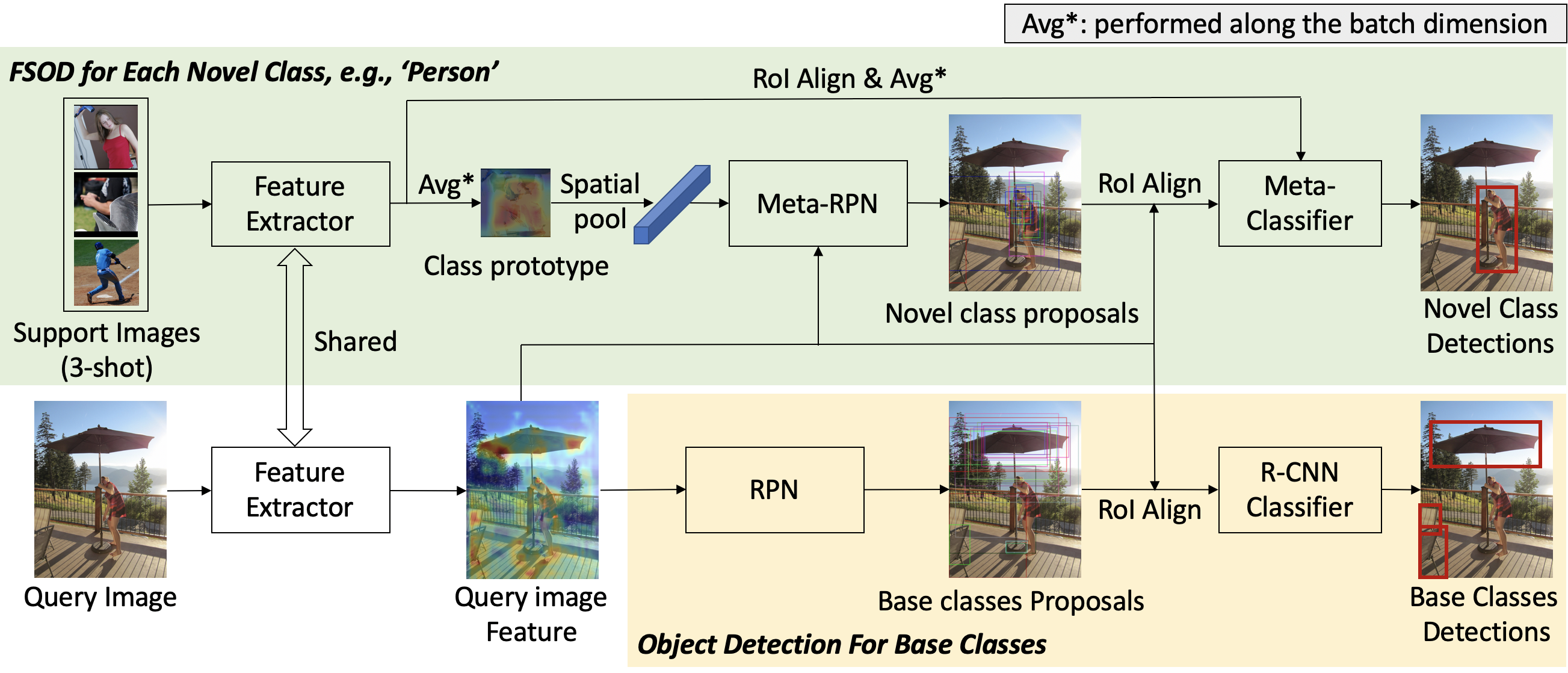}
\end{center}
\caption{Our proposed object detection model for both many-shot base classes and few-shot novel classes. The detection of base and novel classes are decoupled into two branches with a shared feature extractor. For base classes, we follow the original Faster R-CNN detection pipeline. For novel classes, we propose a novel few-shot detector with a two-stage coarse-to-fine prototype matching network. 
Our method can improve the accuracy of few-shot object detection for novel classes, and also maintain high accuracy for base classes. }
\label{detection_base_novel}
\end{figure*}

This has motivated research on few-shot object detection (FSOD) \cite{karlinsky2019repmet}. Given a set of base classes with plenty of examples and another set of novel classes with only few examples, the goal is to transfer the knowledge learned from base classes to novel classes to assist in object detection for novel classes. How to learn few-shot object detectors both efficiently and effectively remains challenging. 

Object proposal \cite{hosang2015makes} is a key ingredient in modern object detectors \cite{ren2015faster,liu2016ssd}, which usually first generate a few potential object bounding boxes (a.k.a proposals) or dense anchor boxes and then convert object detection to a classification task\footnote{We omit the bounding box regression for simplicity}. Training the region proposal network (RPN \cite{ren2015faster}) and proposal classification network (Fast R-CNN \cite{Fast_R-CNN}) using a large scale dataset have been demonstrated successful. However, 
as shown in Table \ref{tab:proposal}, the quality of proposals for few-shot novel classes using existing methods are far worse than that of many-shot base classes. 
Specifically, the detector could miss some high IoU proposals for novel classes due to misclassification with very few examples. Moreover, the spatial location of novel-class proposals usually do not tightly align with the true objects, and may contain only part of the objects, including large background areas, which makes the following few-shot proposal classification challenging.

To address the aforementioned issues, we propose a meta-learning based FSOD method consisting of two modules, in Figure \ref{introduction}. 
\textbf{First}, the Meta-RPN is proposed to generate category-specific proposals for few-shot novel classes both efficiently and effectively. 
Specifically, we use a lightweight non-linear matching network to measure the similarity between the dense sliding windows (a.k.a anchor boxes \cite{ren2015faster}) in the query image feature map and the few-shot novel classes, 
instead of the traditional simple linear object/nonobject classifier, e.g., used in RPN \cite{ren2015faster}, 
thus improving the proposal recall for few-shot novel classes and easing the burden of the following fine-grained proposal classification.
\textbf{Second}, the Meta-Classifier is proposed to measure the similarity between the noisy proposals and few-shot novel classes. 
Our key observation is that performing spatial alignment and focusing on corresponding foreground regions between the high-resolution proposal features and class prototypes \cite{snell2017prototypical} is crucial for few-shot proposal classification. To this end, we propose to estimate soft correspondences between each spatial position in the proposal features and class prototypes. Then based on the soft correspondences, we learn to perform spatial alignment between the two features and discover foreground regions. After that, a non-linear prototype matching network is learned to measure the similarity of the aligned features.

The whole network, denoted as Meta Faster R-CNN, can be trained using meta-learning. After meta-training on the data-abundant base classes, our method can enroll few-shot novel classes \textbf{incrementally} without any training during meta-testing. Our meta-leaning models achieve competitive results compared with the state-of-the-arts (SOTAs) fine-tuned on novel classes. 
With further fine-tuning, we can achieve SOTA accuracy on multiple FSOD benchmarks.

To enable base classes detection in our model, previous methods \cite{wang2020few,wu2020multi,xiao2020few,yan2019meta} usually build a softmax based detector with both base and novel classes. 
We argue that the softmax detector is inflexible to add new classes because it always needs to fine-tune a new classifier.
We also show in Table \ref{tab:base_class} that our few-shot detector is not suitable for base classes considering both running speed and detection accuracy. 
Considering both strengths and weaknesses of the two detectors, we propose to take advantage of the two detectors, by learning a separate Faster R-CNN detection head for base classes using the shared feature backbone, as shown in Figure \ref{detection_base_novel}. Experimental results demonstrate the effectiveness of our model for both base and novel classes.

Our contributions include:
   \textbf{(1)} We propose to learn a coarse-grained prototype matching network (Meta-RPN) with a metric-learning based non-linear classifier to generate class-specific proposals for novel classes with high recall.
   \textbf{(2)} We propose to learn a fine-grained prototype matching network (Meta-Classifier) with a spatial feature alignment and a foreground attention module to address the spatial misalignment issue between proposal features and class prototypes, which improves the overall detection accuracy.
   \textbf{(3)} Considering both strengths and weaknesses of the softmax based detector and our few-shot detector, we propose to take advantage of the two detectors to detect both base and novel classes by learning two detection heads.
   \textbf{(4)} We achieve SOTA results on multiple FSOD benchmarks.

\section{Related Work}

\textbf{Object Detection.} Recently, deep learning \cite{ALEXNET,he2016deep} based methods have dominated the SOTA of object detection. They can be roughly grouped into proposal based methods \cite{ren2015faster,he2017mask,R_RPN,han2018semi} and proposal-free methods \cite{redmon2016you,liu2016ssd,SSD_TDR,tian2019fcos}. Our method belongs to the first kind as our goal is to push the limit of detection accuracy, which is still the top priority for FSOD. 

\noindent \textbf{Few-Shot Object Detection.} 
Building on modern object detectors (e.g., Faster R-CNN \cite{ren2015faster}), existing works \cite{wang2020few,wu2020multi,kang2019few,fan2020few,perez2020incremental,yan2019meta,xiao2020few,hsieh2019one,osokin2020os2d,wang2019meta,wu2020meta,Han_2021_ICCV,Han_2022_CVPR,Han_2022_arXiv} have explored adapting the current detection pipeline to the few-shot setting, including both proposal generation and proposal classification. 
For \textbf{proposal generation,} many current methods \cite{karlinsky2019repmet,wang2020few} directly use the Region Proposal Network (RPN \cite{ren2015faster}) trained on base classes to generate proposals for novel classes. However, it could miss some high IoU boxes for novel classes as novel-classes boxes are regarded as background regions in RPN training over base classes. Fine-tuning RPN on novel classes \cite{wu2020multi,xiao2020few} could improve the performance, but the generalization ability to unseen classes is limited. Other methods \cite{kang2019few,hsieh2019one,fan2020few} propose to modulate query image features with few-shot classes in order to generate category-specific proposals. However, the simple linear object/nonobject classification in RPNs often lacks the robustness in detecting high-quality proposals needed for FSOD.
For \textbf{proposal classification and bbox regression,} few-shot learning methods \cite{vinyals2016matching,snell2017prototypical,sung2018learning,finn2017model,gidaris2018dynamic,Ma_2021_ICCV,Huang_2022_CVPR,ypsilantis2021met}, especially prototypical networks \cite{snell2017prototypical} are introduced to extract prototype representation for each class, and then classification can be performed by using a neural network to measure the similarity between proposal features and class prototypes \cite{sung2018learning,koch2015siamese}. This has been demonstrated effective in FSOD. 
However, they ignore the spatial misalignment issue: similar semantic regions do not appear at the same spatial position between the noisy proposals and few-shot support images.
Our proposed model with Meta-RPN and Meta-Classifier could alleviate the above-mentioned issues.

\section{Our Approach}

\subsection{Task Definition}
\label{task_definition}
In few-shot object detection task, we have two disjoint sets of classes $\mathcal{C} = \mathcal{C}_{base} \cup \mathcal{C}_{novel}$, including base classes $\mathcal{C}_{base}$ and novel classes $\mathcal{C}_{novel}$, and $\mathcal{C}_{base} \cap \mathcal{C}_{novel} = \emptyset$. 

For the base classes, we have plenty of labeled training images $ \mathcal{D} = \{(I, y), I \in \mathcal{I}, y \in \mathcal{Y}\} $ with bounding box annotations, where $I$ is a training image, $y$ is the ground-truth labels for $I$. Specifically, $y=\{ (c_i, box_i), c_i \in \mathcal{C}_{base}, box_i = \{ x_i, y_i, w_i, h_i\} \}_{i=1}^{N}$, containing $N$ bounding boxes in the image $I$ with both class label $c_i$ and box location $box_i$ for each bounding box.

For the novel classes, also known as support classes, we only have $K$-shot (e.g., $K=1,5,10$) labeled samples for each class, also known as support images. Specifically, the support images for novel class $c \in \mathcal{C}_{novel} $ is $\mathcal{S}_c = \{ (I_i^c, box_i), I_i^c \in \mathcal{I}, box_i = \{ x_i, y_i, w_i, h_i\} \}_{i=1}^{K}$, where $I_i^c$ is a training image, $box_i$ is the box location of the object with class label $c$.

The goal is to detect objects of novel classes using few-shot examples and also keep high accuracy for base classes.

\subsection{The Model Architecture}
\label{Model_Architecture}

The key idea of the meta-learning based FSOD is to learn how to match the few-shot classes with query images using the abundant training data of base classes, so that it can generalize to few-shot novel classes. 
Considering the difficulty of object detection with few-shot examples, we propose to learn the detection model via a coarse-to-fine manner following \cite{ren2015faster}.
Meanwhile, we learn a separate Faster R-CNN detection head for base classes using plenty of training samples. 
Our model mainly has the following four modules, as shown in Figure \ref{detection_base_novel}.

\textbf{(1) Feature Extraction.} We use a siamese network to extract features for both query images and the support images. Formally, given a query image $I_q \in \mathbb{R}^{H_q\times W_q\times 3}$, a deep feature backbone is used to extract CNN features $f_q = F(I_q) \in\mathbb{R}^{H \times W \times C}$, where $H$, $W$, $C$ are the height, width and channel dimension of the extracted features. Typically we use the output after $res4$ block in ResNet-50/101 \cite{he2016deep} as our default image features.

For the support images, we first extend the original object bounding boxes with some surrounding context regions as the common practice in previous works \cite{fan2020few,kang2019few,yan2019meta,xiao2020few}, and crop the corresponding object regions in the image. 
Then the cropped images are adjusted to the same size \cite{fan2020few}, and fed into the shared feature backbone to extract the CNN features $F(I_i^c)$ for each support image $I_i^c$. 

\textbf{(2) Object Detection for Base Classes.} 
On top of the feature extraction network, RPN is used to generate category-agnostic proposals of all base classes in the image. 
After that, for each proposal, an R-CNN classifier \cite{Fast_R-CNN} is employed to produce softmax probabilities over all base classes $\mathcal{C}_{base}$ plus a “background” class and bbox regression.

\begin{figure}[t]
\begin{center}
\includegraphics[scale=0.31]{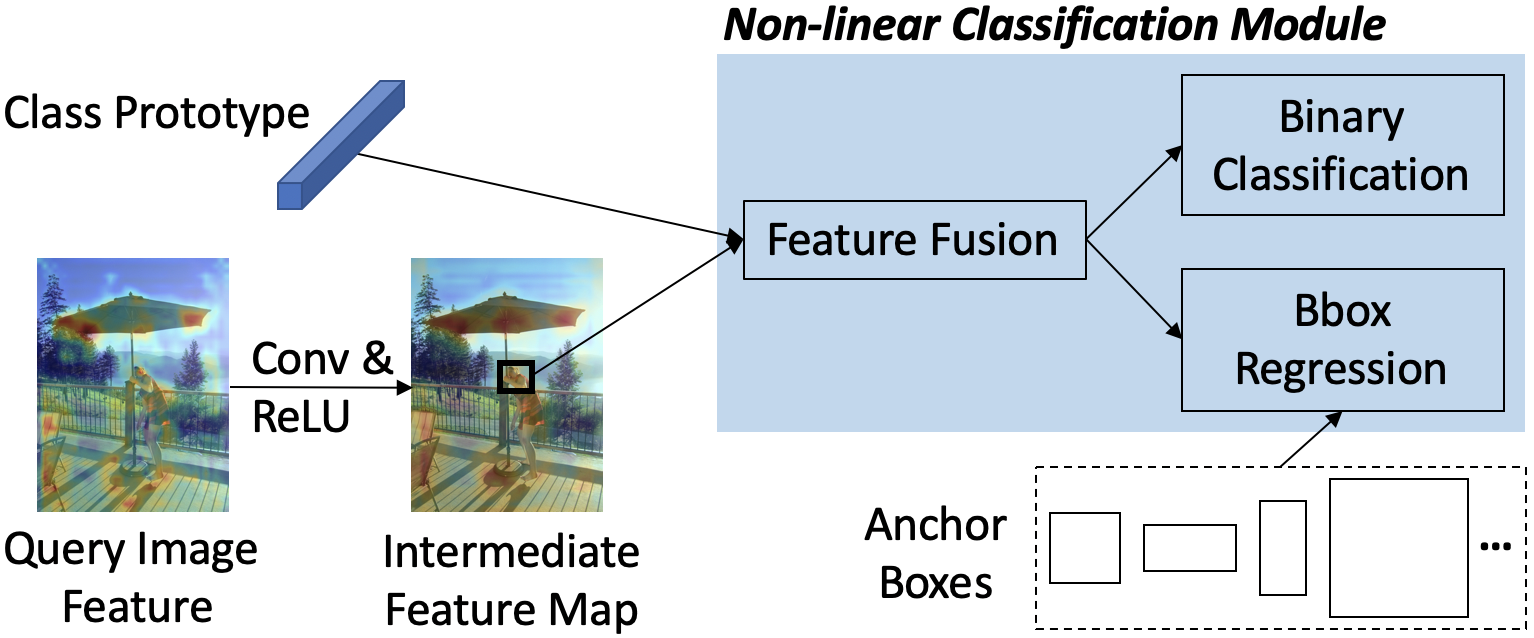}
\end{center}
\caption{Our proposed Meta-RPN with a non-linear classification module.} 
\label{meta_rpn}
\end{figure}

\textbf{(3) Proposal Generation for Novel Classes.} We aim to generate proposals for few-shot novel classes both efficiently and effectively.
In Figure \ref{meta_rpn}, similar to RPN, we attach a small subnetwork on top of $f_q$ to generate proposals centered at each spatial position of $f_q$. Specifically, a $3 \times 3$ conv and ReLU layer is used to extract features for multi-scale anchors centered at each spatial position.
For each novel class, We take the averaged CNN features of the $K$-shot support images as the class prototype $\{{f^c}=\frac{1}{K}\sum_{i=1}^{K}F(I_i^c), c \in \mathcal{C}_{novel}\}$. Then in order to get the same feature size as the anchor boxes, we conduct spatial average pooling to get the pooled prototype $f^c_{pool} = \frac{1}{H*W}\sum_{h,w} f^c$ for each novel class, such that $f^c_{pool} \in \mathbb{R}^C$ captures global representation of class $c$.

Then intead of using a \textbf{simple linear object/nonobject classifier in RPN}, we propose to build a \textbf{metric-learning based non-linear classifier} with a lightweight feature fusion network to better calculate the similarity between the class prototypes and anchor features.
Based on previous work \cite{fan2020few,xiao2020few}, we propose a stronger feature fusion network with Multiplication (Mult), Subtraction (Sub), and Concatenation (Cat) subnetworks. In the first two subnetworks, we perform the basic element-wise fusion operations. Mult can highlight relevant and common features, and Sub can directly measure the distance between two inputs. Cat can be seen as a learnable operation. Although Sub can be learned by Cat and a convolutional layer, we show by experiments that directly learning a Cat subnetwork is difficult while the other two subnetworks can act as shortcuts, and all the three subnetworks are complementary. Formally, 
\begin{equation}
f_q^c = [\Phi_{Mult}(f_q \odot f^c_{pool}), \Phi_{Sub}(f_q-f^c_{pool}), \Phi_{Cat}[f_q, f^c_{pool}]]
\end{equation}
where $\Phi_{Mult}$, $\Phi_{Sub}$ and $\Phi_{Cat}$ all consist of a conv and ReLU layer, and $[,]$ denotes channel-wise concatenation. 
Then $f_q^c$ is fed into a binary classification and bbox regression layer to predict proposals. The proposed feature fusion network can be naturally implemented with convolutional layers, and is computationally efficient and achieve high recall for novel classes using a few proposals.

\begin{figure}[t]
\begin{center}
\includegraphics[scale=0.298]{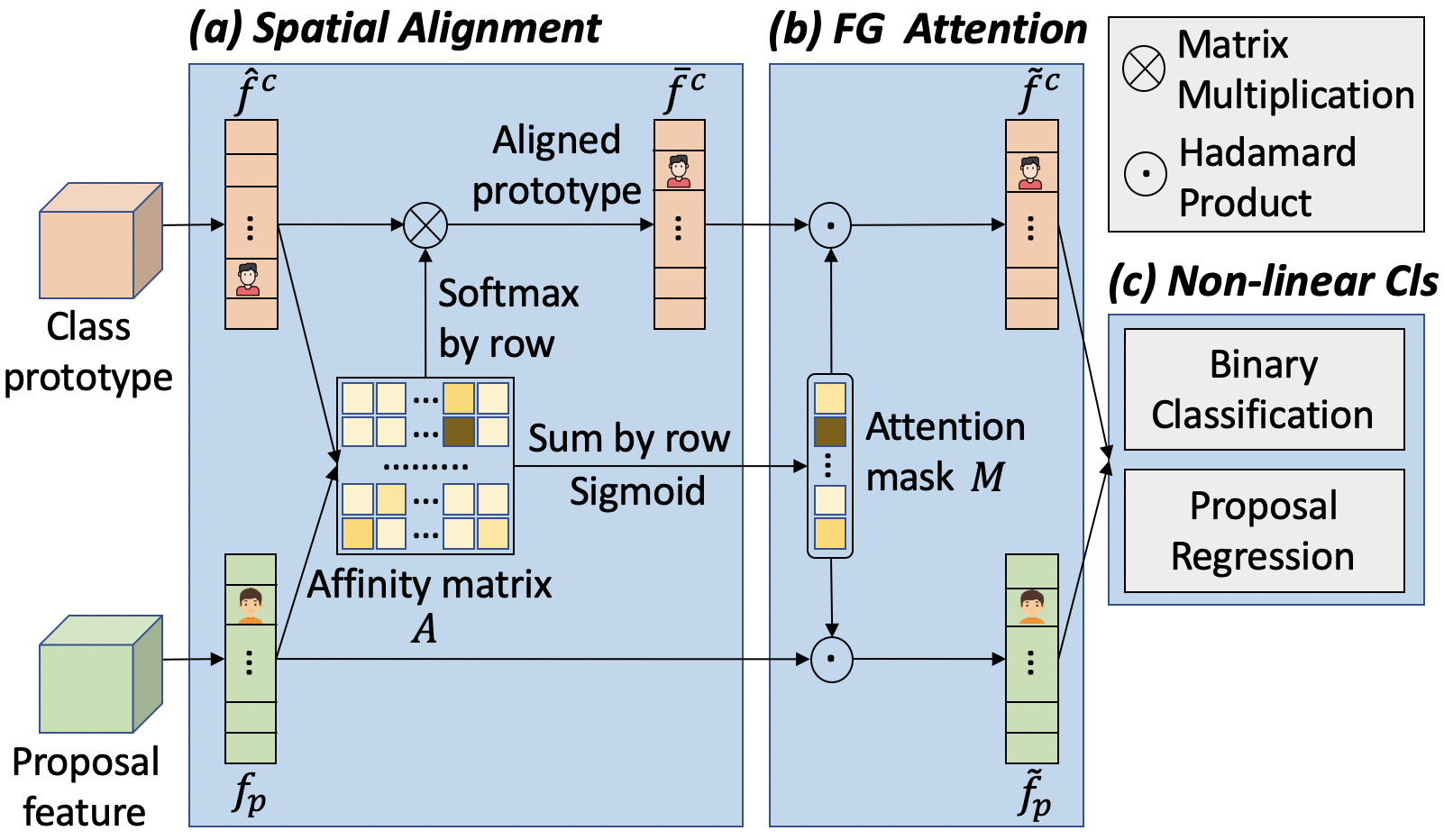}
\end{center}
\caption{Our proposed Meta-Classifier with Attentive Feature Alignment. To calculate the similarity between the class prototype and proposal feature, we first calculate the proposal-aligned class prototype using the spatial alignment module. Then an attention module is proposed to discover foreground regions. Finally, a non-linear classification module is followed to calculate the similarity.}
\label{meta_classifier}
\end{figure}

\textbf{(4) Proposal Classification and Refinement for Novel Classes.}
To calculate the similarity of the generated proposals with novel classes, we first use a siamese network (including RoIAlign \cite{he2017mask} and the $res5$ block in ResNet) to extract proposal features $f_p$ and class prototype $\hat{f}^c$ of the same size $\mathbb{R}^{H'\times W'\times C'}$, where $H'=W'=7, C'=2048$. Different from using the spatially-pooled class prototypes in Meta-RPN, we use high-resolution features for fine-grained matching thanks to the cascade design. However, due to the inaccurate localization of proposals for novel classes, the \textbf{spatial misalignment} between proposals and class prototypes have a negative effect on the few-shot classification.

We propose an attention based feature alignment method to address the spatial misalignment issue. As illustrated in Figure \ref{meta_classifier}, we first establish soft correspondences between two input features by calculating the affinity matrix. The affinity matrix is then used to compute proposal-aligned prototype and localize foreground regions. A non-linear classification module is then followed to calculate the similarity score. 

\textbf{(a) Spatial Alignment Module.} Formally, given a pair of input features $f_p$ and $\hat{f}^c$, both feature maps have $H'\times W'$ CNN grid features, each with a embedding dimension of $\mathbb{R}^{C'}$. We compute the affinity matrix $A \in \mathbb{R}^{H'W'\times H'W'}$, with each item the dot product of two embeddings,
\begin{equation}
A(i,j) = f_p(i) \hat{f}^c(j)^\intercal
\end{equation}
where $A(i,j)$ denotes the similarity of embeddings $f_p(i)$ and $\hat{f}^c(j)$ at spatial location $i$ and $j$ of the proposal feature and class prototype respectively. 
Next is to calculate the proposal-aligned class prototype. The feature alignment is achieved by taking the weighted average of all CNN grid features in the prototype according to the dense semantic correspondence in the affinity matrix $A$. An example is shown in Figure 4. We show the importance of the alignment direction in Table \ref{tab:feature_alignment}.

Formally, for a spatial location $i \in \mathbb{R}^{H'\times W'}$ in $f_p$, softmax normalization is performed over all spatial locations in $ \hat{f}^c(j)$,
\begin{equation}
A(i,j)^{'} = \frac{\exp(A(i,j))}{\sum_{k} \exp(A(i,k))}
\end{equation}
Then, the proposal-aligned class prototype $\bar{f}^c$ at each spatial location $i\in \mathbb{R}^{H'\times W'}$, can be calculated by aggregating the embeddings of all positions in class prototype $\hat{f}^c$ using the normalized similarity,
\begin{equation}
\bar{f}^c(i)=\sum_j A(i,j)^{'}\hat{f}^c(j)
\end{equation}

\textbf{(b) Foreground Attention Module.} As the proposals may contain undesired background regions, a foreground attention mask $M\in\mathbb{R}^{H'\times W'}$ is generated to highlight the corresponding object regions. Formally, for each spatial location $i$ in $f_p$, we summarize the similarity of $f_p(i)$ to each spatial location in $\hat{f}^c$ using the affinity matrix $A$. Then sigmoid function ($\sigma$) is applied to get the normalized probability,
\begin{equation}
M(i) = \sigma(\sum_{j} A(i,j))= \frac{1}{1+\exp(-\sum_{j} A(i,j))}
\end{equation}
where higher values in $M$ indicate that the corresponding locations in $f_p$ are more similar to that of the aligned prototype $\bar{f}^c$, and are more likely to be the same semantic parts. On the other hand, the background regions in the proposals can hardly find corresponding locations in the class prototypes with high similarity, leading to lower values in $M$. 
Therefore we multiply the attention mask $M$ with both $f_p$ and $\bar{f}^c$ to focus on the corresponding foreground regions,
\begin{equation}
\grave{f}^c = M\odot\bar{f}^c, \;\;\; \grave{f}_p = M\odot f_p
\end{equation}

In addition, we further multiply the features $\grave{f}^c$ and $\grave{f}_p$ with learnable parameters $\gamma_1$ and $\gamma_2$ (both initialized as $0$) and add back the input features for stable training,
\begin{equation}
\widetilde{f}^c = \gamma_1\grave{f}^c+\hat{f}^c, \;\;\; \widetilde{f}_p = \gamma_2\grave{f}_p + f_p
\end{equation}

\textbf{(c) Non-linear Classification Module.} To measure the similarity between the final features $\widetilde{f}^c$ and $\widetilde{f}_p$, we employ a feature fusion network to aggregate the two features with high-resolution,
\begin{equation}
f = [\Psi_{Mult}(\widetilde{f}^c \odot \widetilde{f}_p), \Psi_{Sub}(\widetilde{f}^c-\widetilde{f}_p), \Psi_{Cat}[\widetilde{f}^c, \widetilde{f}_p]]
\end{equation}
where $\Psi_{Mult}$, $\Psi_{Sub}$ and $\Psi_{Cat}$ are three similar CNN networks, each with 3 conv and ReLU layers for non-linear fusion. After that, a binary classification and bbox regression layer are followed for final detection.

\subsection{The Training Framework}
\label{Training_Framework}

Our training framework has the following three steps,

\textbf{Meta-learning with base classes.} We learn our Meta-RPN and Meta-Classifier using meta-training. In each training episode, we sample a few classes from base classes, each class with $k$-shot support images, to simulate the FSOD scenario for novel classes. Besides, we also sample a few query images with ground-truth boxes, and use binary cross-entropy loss and smooth $L1$ loss \cite{Fast_R-CNN} for model training. To prevent the vast number of negative matching pairs from overwhelming the training loss, we keep a ratio of 1:3 for positive and negative matching pairs for balanced model training.
After meta-training, the model can be directly applied to novel classes during meta-testing without any training.

\textbf{Learning the separate detection head for base classes.} After meta-training, we fix the parameters of the backbone feature extractor, and learn the RPN and R-CNN module for base classes following \cite{ren2015faster}.

\textbf{Fine-tuning with both base and novel classes.} We only use base-classes dataset for training in the first two step. For fine-tuning, we sample a small balanced dataset (original images) of both base and novel classes following the common practice in previous works \cite{kang2019few,yan2019meta,xiao2020few,fan2020few,wu2020multi,wang2020few}. We make sure that the total number instances for each novel class is exactly $k$-shot in the sampled dataset, which are also used as the support set during testing.

The key difference of meta-learning and fine-tuning is that, there is no training on novel classes in meta-learning. We only use the support set of novel classes to calculate prototypes during meta-testing. The support images are cropped from the original images using ground-truth annotations.
While during fine-tuning, we use the original novel-classes images as query images to fine-tune our few-shot detector, including both the Meta-RPN and Meta-Classifier. The model performance for novel classes will improve when we gradually use more images for fine-tuning.

\section{Experimental Results}
\label{Experimental_Results}

\subsection{Datasets}
\label{datasets}

We use two widely-used FSOD benchmarks MSCOCO \cite{lin2014microsoft} and PASCAL VOC \cite{everingham2010pascal} for model evaluation, and follow FSOD settings the same as previous works \cite{kang2019few,wang2020few} by using the exact same few-shot images for fair comparison. 

More implementation details are included in the supplementary material.

\begin{table*}[ht]
\centering
\footnotesize
\setlength{\tabcolsep}{0.4em}
\adjustbox{width=\linewidth}{
\begin{tabular}{l|ccc|ccc|ccc|ccc|ccc}
\toprule
&\multicolumn{3}{c|}{Proposal Generation} &\multicolumn{3}{c|}{Proposal Classification}  &\multicolumn{3}{c|}{2-shot} & \multicolumn{3}{c|}{10-shot} & \multicolumn{3}{c}{30-shot}\\
& RPN &Attention-RPN$^\dagger$ & Meta-RPN & NL-Cls$^\ddagger$ & Alignment & FG-Attention & AP & AP50 & AP75 & AP & AP50 & AP75 & AP & AP50 & AP75 \\ \midrule
\multicolumn{16}{c}{\textbf{Meta-training the model on base classes, and meta-testing on novel classes}} \\ \midrule
(a) & \checkmark & &  & \checkmark (LR$^\S$) &  & & 4.9 & 10.5 & 4.1 &   6.8 & 13.5 & 6.0   &  7.6 & 15.4 & 6.6 \\
(b) & \checkmark & &  & \checkmark & & & 5.1 & 11.0 & 4.3 &   7.2 & 14.5 & 6.4  & 8.3 & 16.9 & 7.4 \\
(c) & & \checkmark &  & \checkmark & & & 5.5 & 11.6 & 4.7 &   7.7 & 15.7 & 6.9  & 9.0 & 18.5 & 7.8 \\
(d) & & & \checkmark & \checkmark & & & 6.0 & 12.7 & 5.0 &  8.3 & 16.7 & 7.3 & 9.5 & 19.1 & 8.4 \\
(e) & & & \checkmark & \checkmark & \checkmark & & 6.4 & 13.6 & 5.4 & 8.6 & 17.5 & 7.6 & 9.8 & 20.0 & 8.7 \\
(f) & & & \checkmark & \checkmark & \checkmark & \checkmark & {7.0} & {14.7} & 5.9 & 9.2 & 18.3 & 8.0 & 10.2 & 20.5 & 9.3 \\
(g)$^\P$ & & & \checkmark & \checkmark & \checkmark & \checkmark & {7.0} & 13.5 & \textbf{6.4} & {9.7} & {18.5} & {9.0} & {11.3} & {21.2} & {10.6} \\ \midrule
\multicolumn{16}{c}{\textbf{Fine-tuning the model on novel classes, and testing on novel classes}} \\ \midrule
(h) & & & \checkmark & \checkmark & \checkmark & \checkmark & 7.0 & 15.3 & 5.7 & 11.1 & 23.4 & 9.5 & 15.4 & 31.7 & 13.4 \\ 
(i)$^\P$ & & & \checkmark & \checkmark & \checkmark & \checkmark & \textbf{7.6} & \textbf{16.3} & {6.2} & \textbf{12.7} & \textbf{25.7} & \textbf{10.8} & \textbf{16.6} & \textbf{31.8} & \textbf{15.8} \\
\bottomrule
\end{tabular}}
\caption{{Ablation study on each component in our approach. We show results of the novel classes on the MSCOCO dataset under shots 2, 10 and 30.}  $^\dagger$ \cite{fan2020few}. $^\ddagger$ Our Non-Linear Classifier. $^\S$ Using Low-Resolution features. $^\P$ Using ResNet-101, otherwise ResNet-50.} 
\label{tab:ablation}
\end{table*}

\subsection{Ablation Study}
\label{Ablation_Study}

\textbf{Effectiveness of our Meta-RPN.} We compare three different proposal generation methods for novel classes (RPN, Attention-RPN \cite{fan2020few}, and our Meta-RPN) in Table \ref{tab:ablation} (b), (c) and (d) and Table \ref{tab:proposal}. \textbf{(1)} The object/nonobject classifier in RPN, pretrained on base classes, do not generalize well to novel classes as only base-classes regions are regarded as true objects in training and it may miss some high IoU boxes for novel classes due to misclassification. 
Both proposal AR and detection AP of novel classes using RPN are much lower than the others. 
\textbf{(2)} Using class prototype to reweight query image features in Attention-RPN \cite{fan2020few} can generate class-specific proposals and improve both proposal AR and detection AP. 
\textbf{(3)} By using a metric-learning based non-linear classifier with a lightweight feature fusion network, our Meta-RPN can perform better prototype matching compared with the simple linear object/nonobject classifier used in Attention-RPN \cite{fan2020few}.
Our Meta-RPN consistently improves using different number of proposals, especially after fine-tuning. We use 100 proposals for each novel class by default.

\begin{table}[t]
    \centering
    \footnotesize
    \adjustbox{width=\linewidth}{
    \begin{tabular}{l|ccc}
    \toprule
    {\#Proposals$^\dagger$} & {RPN} & {Attention-RPN$^\ddagger$} & {Our Meta-RPN} \\ \midrule
    \multicolumn{4}{c}{\textbf{Meta-training the model on base classes,}} \\
    \multicolumn{4}{c}{\textbf{and meta-testing on novel classes}} \\ \midrule
    10    & {7.8}     & 18.0    & \textbf{18.3} \\
    100  & {20.5}    & 32.7    & \textbf{33.2} \\
    1000 & {35.0}    & 44.1    & \textbf{44.9} \\ \midrule

    \multicolumn{4}{c}{\textbf{Fine-tuning the model on novel classes,}} \\
    \multicolumn{4}{c}{\textbf{and testing on novel classes}} \\ \midrule
    10    & 9.2    & 18.7    & \textbf{18.9} \\
    100  & 22.9   & 33.2    & \textbf{33.8} \\
    1000 & 37.4   & 44.7    & \textbf{45.8} \\ \midrule
    \multicolumn{4}{c}{\textbf{Base classes: AR@10=27.6, AR@100=43.3, AR@1000=51.0}} \\
    \bottomrule
    \end{tabular}}
    \caption{{Proposal average recall (AR) using the models in Table \ref{tab:ablation} (b), (c) and (d), with 10-shot meta-testing and 10-shot fine-tuning. $^\dagger$ The number of proposals generated for each novel class. $^\ddagger$ \cite{fan2020few}.} } 
\label{tab:proposal}
\end{table}

\begin{figure}[t]
\begin{center}
\includegraphics[scale=0.25]{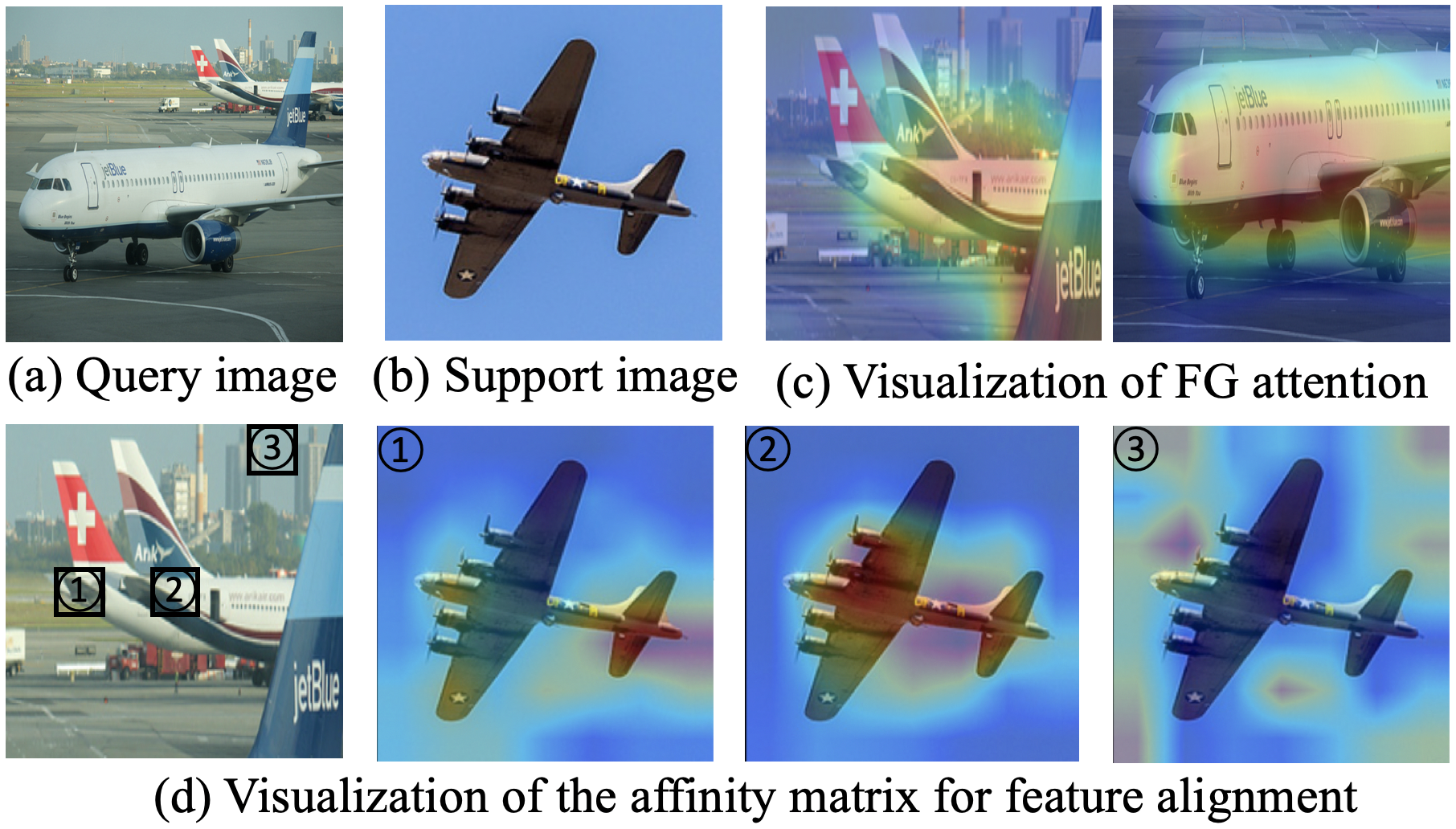}
\end{center}
\caption{Visualization of the affinity matrix for alignment and foreground attention masks. In (d), we use 3 black boxes in the proposal as query, and show on the right the corresponding similarity maps in the 1-shot support image.}
\label{visualization}
\end{figure}

\textbf{Effectiveness of our Meta-Classifier.} \textbf{(1)} We first show that using high-resolution features in our Meta-Classifier can achieve good results by comparing Table \ref{tab:ablation} (a) and (b). This is because fine-grained details could provide important clues when performing matching between two features. \textbf{(2)} Although using high-resolution feature maps improve matching performance, the large spatial misalignment between proposals and few-shot classes is harmful for matching due to the inaccurate proposal localization. Using our attentive feature alignment with both spatial alignment and foreground attention module, the performance can be improved consistently for most of the shots and metrics by comparing models in Table \ref{tab:ablation} (d), (e) and (f).
\textbf{(3)} We show in Table \ref{tab:feature_alignment} the importance of the alignment direction. We can keep the original structural information in the proposal for precise bbox regression, when aligning the class prototype to the proposal feature.
\textbf{(4)} We also show that using a stronger feature backbone ResNet-101 can have better results than ResNet-50 in Table \ref{tab:ablation} (f-g) and (h-i), especially after fine-tuning. 
\textbf{(5)} Visualization of the affinity matrix and attention masks are shown in Figure \ref{visualization}.

\begin{table}[t]
    \centering
    \footnotesize
    \addtolength{\tabcolsep}{-1pt}
    \adjustbox{width=\linewidth}{
    \begin{tabular}{ccc|ccc|ccc}
    \toprule
    \multirow{2}{*}{Mult} & \multirow{2}{*}{Sub} & \multirow{2}{*}{Cat} 
    & \multicolumn{3}{c|}{Meta-RPN} & \multicolumn{3}{c}{Meta-Classifier}\\
    & & & AP & AP50 & AP75 & AP & AP50 & AP75 \\ \midrule
    \checkmark & & & 7.7 & 16.0 & 6.9  & 7.3 & 15.9 & 6.1 \\
    & \checkmark & & 7.6 & 15.8 & 6.7  & 4.9 & 10.2 & 4.5 \\
    & & \checkmark & 7.4 & 15.3 & 6.6  & 5.3 & 10.4 & 5.2 \\
    \checkmark & \checkmark & & 8.0 & 16.3 & 7.1  & 7.6 & 16.3 & 6.7 \\
    \checkmark & & \checkmark & 8.1 & 16.5 & 7.2  & 7.9 & 16.5 & 7.1 \\
    & \checkmark & \checkmark & 7.8 & 16.1 & 6.9  & 5.7 & 10.9 & 5.4 \\ \midrule
    \checkmark & \checkmark & \checkmark & \textbf{8.3} & \textbf{16.7} & \textbf{7.3}  & \textbf{8.3} & \textbf{16.7} & \textbf{7.3} \\
    \bottomrule
    \end{tabular}}
    \caption{Ablation study on feature fusion, using model in Table 1(d) with only meta-training and 10-shot meta-testing.} 
\label{tab:feature_fusion}
\end{table}

\begin{table}[t]
    \centering
    \footnotesize
    \adjustbox{width=0.95\linewidth}{
    \begin{tabular}{c|ccc|ccc}
    \toprule
    \multirow{2}{*}{Direction}
    & \multicolumn{3}{c|}{2-shot} & \multicolumn{3}{c}{10-shot}\\
    & AP & AP50 & AP75 & AP & AP50 & AP75 \\ \midrule
    $\hat{f}^c \leftarrow f_p$ & {5.1} & {11.2} & {4.0}   & {6.5} & {13.5} & {5.7} \\
    $\hat{f}^c \rightarrow f_p$ & \textbf{7.0} & \textbf{14.7} & \textbf{5.9} & \textbf{9.2} & \textbf{18.3} & \textbf{8.0} \\
    \bottomrule
    \end{tabular}}
    \caption{Ablation study on the direction of feature alignment, using model in Table 1(f) with only meta-training. $\hat{f}^c$ is the novel class prototype, and $f_p$ is the proposal features.} 
\label{tab:feature_alignment}
\end{table}

\textbf{Effectiveness of our non-linear feature fusion network.} As shown in Table \ref{tab:feature_fusion}, we perform ablation study of our proposed feature fusion network in both Meta-RPN and Meta-Classifier. We can find that using simple element-wise fusion operations especially the Mult subnetwork shows good results, which is widely used in previous work \cite{kang2019few,yan2019meta,xiao2020few,fan2020few}. Directly using the Cat subnetwork does not achieve good results in both Meta-RPN and Meta-Classifier. This is because the Cat subnetwork attempts to learn complex fusion between the two features, which is not easy for training and generalization. When combining all three subnetworks together, it could ease the training of the Cat subnetwork and learn complementary fusion beyond the Mul and Sub subnetworks.

\textbf{Effectiveness of both meta-learning and fine-tuning.}
We show the comparison of meta-training and fine-tuning in Table \ref{tab:ablation} (f) and (h) for ResNet-50, and in Table \ref{tab:ablation} (g) and (i) for ResNet-101. We can find that using fine-tuning, the performance improves for large shot settings (e.g., 10/30 shot). However, for extremely few-shot setting (e.g., 2 shot), the performance could hardly improve because fine-tuning is prone to overfitting using very few examples.

\begin{table*}[ht]
\centering
\footnotesize
\addtolength{\tabcolsep}{-3.7pt}
\adjustbox{width=\linewidth}{
\begin{tabular}{l|c|ccccc|ccccc|ccccc}
\toprule
\multirow{2}{*}{Method} & \multirow{2}{*}{Venue} & \multicolumn{5}{c|}{Novel Set 1} & \multicolumn{5}{c|}{Novel Set 2} & \multicolumn{5}{c}{Novel Set 3} \\ 
&  & 1     & 2     & 3    & 5    & 10   & 1     & 2     & 3    & 5    & 10   & 1     & 2     & 3    & 5    & 10   \\ \midrule
\multicolumn{17}{c}{\textbf{Meta-training the model on base classes, and meta-testing on novel classes}} \\ \midrule
Fan et al.\ \cite{fan2020few}$^{\dag}$ & CVPR 2020 & 32.4 & 22.1 & 23.1 & 31.7 & 35.7    & 14.8 & 18.1 & 24.4 & 18.6 & 19.5 &    25.8 & 20.9 & 23.9 & 27.8 & 29.0 \\
Meta Faster R-CNN (Ours) & This work & \textbf{40.2} & \textbf{30.5} & \textbf{33.3} & \textbf{42.3} & \textbf{46.9}   & \textbf{26.8} & \textbf{32.0} & \textbf{39.0} & \textbf{37.7} & \textbf{37.4}   & \textbf{34.0} & \textbf{32.5} & \textbf{34.4} & \textbf{42.7} & \textbf{44.3} \\ \midrule
\multicolumn{17}{c}{\textbf{Fine-tuning the model on novel classes, and testing on novel classes}} \\ \midrule
FSRW~\cite{kang2019few}  & ICCV 2019 & 14.8  & 15.5  & 26.7 & 33.9 & 47.2 & 15.7  & 15.3  & 22.7 & 30.1 & 40.5 & 21.3  & 25.6  & 28.4 & 42.8 & 45.9 \\ 
MetaDet~\cite{wang2019meta} & ICCV 2019 & 18.9 & 20.6 & 30.2 & 36.8 & 49.6 & 21.8 & 23.1 & 27.8 & 31.7 & 43.0 & 20.6 & 23.9 & 29.4 & 43.9 & 44.1 \\ 
Meta R-CNN~\cite{yan2019meta} & ICCV 2019 & 19.9 & 25.5 & 35.0 & 45.7 & 51.5 & 10.4 & 19.4 & 29.6 & 34.8 & 45.4 & 14.3 & 18.2 & 27.5 & 41.2 & 48.1 \\ 
TFA w/ fc \cite{wang2020few} & ICML 2020 & {36.8} & {29.1} & {43.6} & {55.7} & {57.0} & {18.2} & {29.0} & {33.4} & {35.5} & {39.0} & {27.7} & {33.6} & {42.5} & {48.7} & {50.2}\\
TFA w/ cos \cite{wang2020few} & ICML 2020 & 39.8 & 36.1 & 44.7 & 55.7 & 56.0 & 23.5 & 26.9 & 34.1 & 35.1 & 39.1 & 30.8 & 34.8 & 42.8 & 49.5 & 49.8 \\ 
Xiao et al. \cite{xiao2020few} & ECCV 2020 & 24.2 & 35.3 &  42.2 &  49.1 &  57.4 & 21.6 & 24.6 &  31.9 &  37.0 &  45.7 & 21.2 &  30.0 &  37.2 &  43.8 &  49.6 \\
MPSR \cite{wu2020multi} & ECCV 2020 & 41.7 & 42.5 & 51.4 & 55.2 & 61.8 & 24.4 & 29.3 & 39.2 & 39.9 & 47.8 & 35.6 & 41.8 & 42.3 & 48.0 & 49.7 \\ 
Fan et al. \cite{fan2020few}$^{\dag}$ & CVPR 2020 & 37.8 & 43.6 & 51.6 & 56.5 & 58.6    & 22.5 & 30.6 & 40.7 & 43.1 & 47.6    & 31.0 & 37.9 & 43.7 & 51.3 & 49.8\\ 
SRR-FSD \cite{Zhu_2021_CVPR} & CVPR 2021 & \textbf{47.8} & 50.5 & 51.3 & 55.2 & 56.8    & \textbf{32.5} & 35.3 & 39.1 & 40.8 & 43.8    & 40.1 & 41.5 & 44.3 & 46.9 & 46.4 \\
TFA + Halluc \cite{Zhang_2021_CVPR} & CVPR 2021 & 45.1 & 44.0 & 44.7 & 55.0 & 55.9   & 23.2 & 27.5 & 35.1 & 34.9 & 39.0   & 30.5 & 35.1 & 41.4 & 49.0 & 49.3 \\
CoRPNs + Halluc \cite{Zhang_2021_CVPR} & CVPR 2021 & 47.0 & 44.9 & 46.5 & 54.7 & 54.7   & 26.3 & 31.8 & 37.4 & 37.4 & 41.2   & 40.4 & 42.1 & 43.3 & 51.4 & 49.6 \\
FSCE \cite{Sun_2021_CVPR} & CVPR 2021 & 44.2 & 43.8 & 51.4 & 61.9 & 63.4    & 27.3 & 29.5 & 43.5 & 44.2 & 50.2    & 37.2 & 41.9 & 47.5 & 54.6 & 58.5 \\
Meta Faster R-CNN (Ours) & This work & 43.0 & \textbf{54.5} & \textbf{60.6} & \textbf{66.1} & \textbf{65.4}   & 27.7 & \textbf{35.5} & \textbf{46.1} & \textbf{47.8} & \textbf{51.4}   & \textbf{40.6} & \textbf{46.4} & \textbf{53.4} & \textbf{59.9} & \textbf{58.6}\\ 
\bottomrule
\end{tabular}}
\caption{{Few-shot object detection performance (AP50) on the PASCAL VOC dataset.} We use ResNet-101 following most of the previous works. $^{\dag}$ Our reimplementation results.} 
\label{tab:main_voc}
\end{table*}

\begin{table}[t]
    \centering
    \footnotesize
    \adjustbox{width=\linewidth}{
    \begin{tabular}{c|ccc|c}
    \toprule
    \multirow{2}{*}{Model} & \multicolumn{3}{c|}{Accuracy} & {Speed} \\
    & AP & AP50 & AP75 & (/img) \\ \midrule
    Base model & \textbf{39.2} & \textbf{59.3} & \textbf{42.8} & 0.21s \\
    TFA \cite{wang2020few} & 35.8 & 55.5 & 39.4 & 0.21s \\ \midrule
    2-shot meta-testing & 18.5 & 29.4 & 19.8 & 2.1s \\
    10-shot meta-testing & 21.4 & 33.6 & 23.1 & 2.1s \\
    30-shot meta-testing & 22.7 & 35.3 & 24.7 & 2.1s \\ \midrule
    Our model & 36.9 & 57.0 & 40.2 & 0.21s \\
    \bottomrule
    \end{tabular}}
    \caption{{Base classes evaluation results on MSCOCO using ResNet-101.} We compare our final model with the pre-trained base model using the softmax classifier, TFA \cite{wang2020few} 30-shot fine-tuning, and 2/10/30-shot meta-testing using the few-shot model in Table \ref{tab:ablation} (g).} 
\label{tab:base_class}
\end{table}

\begin{figure}
\begin{center}
\includegraphics[scale=0.21]{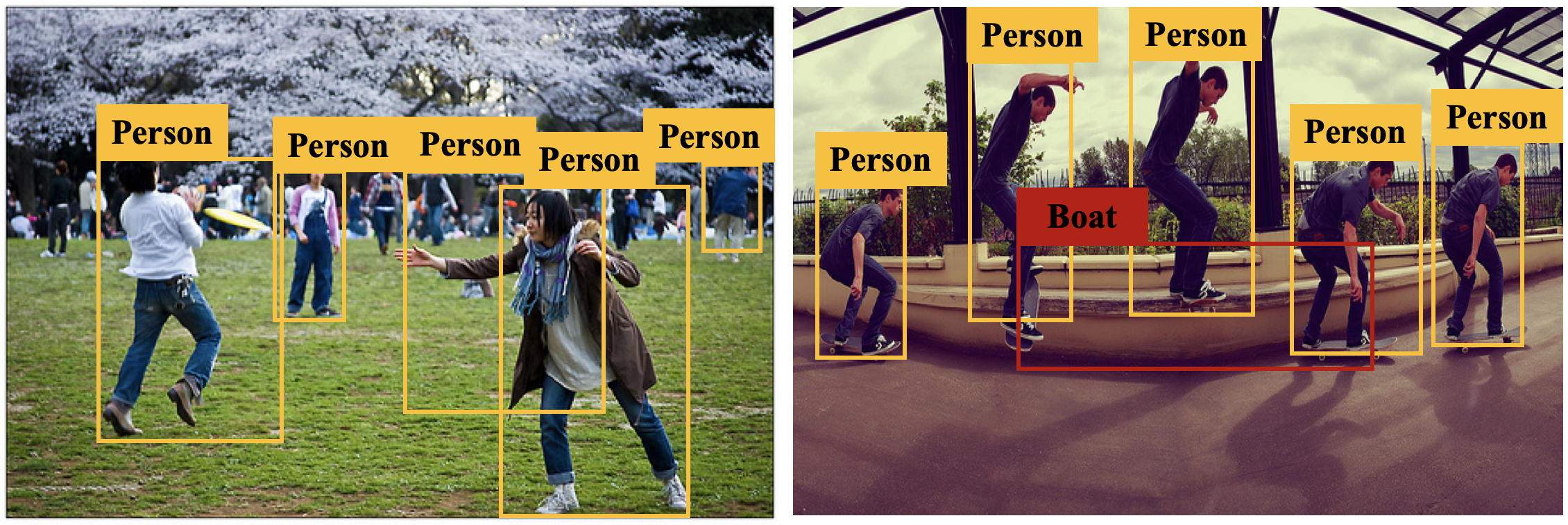}
\end{center}
\caption{Failure cases on COCO dataset. Detection score threshold is 0.5. Left: Missing detection of small objects `Person'. Right: Misclassification of the `Boat'. Future work could improve multi-scale object detection and introduce image context information for detection.}
\label{failure_cases}
\end{figure}

\begin{table*}[ht]
\centering
\footnotesize
\addtolength{\tabcolsep}{-2.5pt}
\adjustbox{width=\linewidth}{
\begin{tabular}{l|c|ccc|ccc|ccc|ccc}
\toprule
\multirow{2}{*}{Model}
& \multirow{2}{*}{Venue} & \multicolumn{3}{c|}{1-shot} & \multicolumn{3}{c|}{2-shot} & \multicolumn{3}{c|}{10-shot} & \multicolumn{3}{c}{30-shot}\\
& & AP & AP50 & AP75 & AP & AP50 & AP75 & AP & AP50 & AP75 & AP & AP50 & AP75 \\ \midrule
\multicolumn{14}{c}{\textbf{Meta-training the model on base classes, and meta-testing on novel classes}} \\ \midrule
Fan et al. \cite{fan2020few}$^{\dag}$ & CVPR 2020 & 4.0 & 8.5 & 3.5  & 5.4 & 11.6 & 4.6    & 7.6 & 15.4 & 6.8  & 8.9 & 17.8 & 8.0 \\
Meta Faster R-CNN (Ours) & This work & \textbf{5.0} & \textbf{10.2} & \textbf{4.6}    & \textbf{7.0} & \textbf{13.5} & \textbf{6.4}      & \textbf{9.7} & \textbf{18.5} & \textbf{9.0} & \textbf{11.3} & \textbf{21.2} & \textbf{10.6} \\ \midrule
\multicolumn{14}{c}{\textbf{Fine-tuning the model on novel classes, and testing on novel classes}} \\ \midrule
FSRW\small{~\cite{kang2019few}} & ICCV 2019 & - & - & - & - & - & - & 5.6 & 12.3 & 4.6 & 9.1 & 19.0 & 7.6 \\ 
MetaDet~\cite{wang2019meta}  & ICCV 2019 & - & - & - & - & - & - & 7.1 & 14.6 & 6.1 & 11.3 & 21.7 & 8.1 \\
Meta R-CNN\small{~\cite{yan2019meta}} & ICCV 2019 & - & - & - & - & - & - & 8.7 & 19.1 & 6.6 & 12.4 & 25.3 & 10.8\\
TFA w/ fc \cite{wang2020few} & ICML 2020 & 2.9 & 5.7 & 2.8   & 4.3 & 8.5 & 4.1     & 10.0 & 19.2 & 9.2 & 13.4 & 24.7 & 13.2  \\
TFA w/ cos \cite{wang2020few} & ICML 2020 & 3.4 & 5.8 & 3.8   & 4.6 & 8.3 & 4.8    & 10.0 & 19.1 & 9.3 & 13.7 & 24.9 & 13.4 \\ 
Xiao et al. \cite{xiao2020few}$^{\ddag}$ & ECCV 2020 & 3.2 & 8.9 & 1.4   & 4.9 & 13.3 & 2.3   & 10.7 & 25.6 & 6.5 & 15.9 & 31.7 & 15.1 \\
MPSR \cite{wu2020multi} & ECCV 2020 & 2.3 & 4.1 & 2.3    & 3.5 & 6.3 & 3.4    & 9.8 & 17.9 & 9.7 & 14.1 & 25.4 & 14.2 \\
Fan et al. \cite{fan2020few}$^{\dag}$ & CVPR 2020 & 4.2 & 9.1 & 3.0   & 5.6 & 14.0 & 3.9     & 9.6 & 20.7 & 7.7  & 13.5 & 28.5 & 11.7 \\
SRR-FSD \cite{Zhu_2021_CVPR} & CVPR 2021 & - & - & - & - & - & - & 11.3 & 23.0 & 9.8  & 14.7 & 29.2 & 13.5 \\ 
TFA + Halluc \cite{Zhang_2021_CVPR} & CVPR 2021 & 4.4 & 7.5 & \textbf{4.9}    & 5.6 & 9.9 & 5.9  & - & - & - & - & - & - \\
CoRPNs + Halluc \cite{Zhang_2021_CVPR} & CVPR 2021 & 3.8 & 6.5 & 4.3    & 5.0 & 9.0 & 5.2   & - & - & - & - & - & - \\
FSCE \cite{Sun_2021_CVPR} & CVPR 2021 & - & - & - & - & - & - & 11.9 & - & 10.5   & 16.4 & - & \textbf{16.2} \\
Meta Faster R-CNN (Ours) & This work & \textbf{5.1} & \textbf{10.7} & 4.3  & \textbf{7.6} & \textbf{16.3} & \textbf{6.2}    & \textbf{12.7} & \textbf{25.7} & \textbf{10.8} & \textbf{16.6} & \textbf{31.8} & 15.8 \\
\bottomrule
\end{tabular}}
\caption{{Few-shot object detection performance on the MSCOCO dataset.} We use ResNet-101 following most of the previous works.
    $^{\dag}$Our reimplementation results using the exact same few-shot training instances as \cite{kang2019few,wang2020few}. $^{\ddag}$The authors report these results at https://github.com/YoungXIAO13/FewShotDetection.} 
\label{tab:main_coco}
\end{table*}

We would like to highlight the advantage of our meta-learning method. 
\textbf{(1) Efficient for adaptation to novel classes}. After meta-training on base classes, there is no further training during meta-testing. To enroll novel classes, we only need to calculate their prototype representations via network inference. Fine-tuning needs additional training over the original novel-classes images. 
\textbf{(2) Effective for extremely few-shot scenarios}. Our meta-learning model shows superior performance compared with fine-tuning using very few examples (e.g., 1/2 shot in Table \ref{tab:main_coco}). 

The above two strengths make our meta-learning method promising for realistic applications. We can first perform meta-training using as many base classes as possible. Then the model parameters are fixed. Users can incrementally add new classes to the system using 1-shot support image without forgetting old classes, and without training/fine-tuning. 

\textbf{Effectiveness of learning a separate detection head for base classes.} 
The results are shown in Table \ref{tab:base_class}. 
\textbf{(1)} The pretrained object detection model on base classes provides an approximate upper bound detection accuracy for base classes. \textbf{(2)} We use the learned few-shot object detector for base-classes object detection. Specifically, we select $K$-shot support images for each base class, and perform object detection for base classes via meta-testing. We find that the meta-testing results on base classes are much higher than novel classes, because the model is meta-trained over base classes, but there is still a big gap compared with the upper bound. The reason is that training a linear classifier with softmax in R-CNN, is more efficient to learn class prototypes than simply averaging the $K$-shot support images, when the training data is sufficient. Meanwhile, the running speed is slow due to independent detection of each class. \textbf{(3)} Our method of learning a separate detection head for base classes, shows better results compared with the strong TFA \cite{wang2020few} model. 

We would like to emphasize the difference between our model and previous methods \cite{yan2019meta,xiao2020few}. Although these methods claim to use meta-learning, they still need to train a softmax based detector with both base and novel classes according to their implementation. The limitation of the softmax based detector is the inflexibility to add new classes to the system, because they need to fine-tune a new softmax based detector with new classes. Our method addresses this problem by decoupling the detection of base and novel classes into two detection heads, and learning a class-agnostic FSOD for novel classes.

\subsection{Comparison with State-of-the-arts}
\label{Comparison_SOTAs}

We compare our final model with the state-of-the-arts on two FSOD benchmarks in Table \ref{tab:main_voc} and \ref{tab:main_coco} respectively.
\textbf{(1)} Most of the previous works only report fine-tuning results. We demonstrate that our meta-learning-only model improves significantly compared with the strong baseline model \cite{fan2020few}, and outperforms or at least attains comparable results compared with other SOTAs using fine-tuning, especially in extremely few-shot setting (e.g., 1/2-shot), where fine-tuning is not effective. \textbf{(2)} We achieve SOTA performance in most of the shots and metrics of the two FSOD benchmarks. The exception is that SRR-FSD \cite{Zhu_2021_CVPR} and Halluc \cite{Zhang_2021_CVPR} have better performance on voc split1\&2 1-shot setting due to the introduction of linguistic semantic knowledge and image hallucination, which is complementary to our model. Moreover, our meta-learning-only model shows better results in the more challenging MSCOCO 1/2-shot settings. 
\textbf{(3)} We show the failure cases in Figure \ref{failure_cases}. More results can be found in the supplementary material.

\section{Conclusion}

We propose a novel meta-learning based few-shot object detection model in this paper. 
Our model consists of the following two modules to tackle with the low quality of proposals for few-shot classes.
First, a lightweight coarse-grained prototype matching network is proposed to generate proposals for few-shot classes in an efficient and effective manner. 
Then, a fine-grained prototype matching network with attentive feature alignment is proposed to address the spatial misalignment between the noisy proposals and few-shot classes.
Experiments on multiple FSOD benchmarks demonstrate the effectiveness of our approach.

\section{Acknowledgments}
This research is based upon work supported by the Intelligence
Advanced Research Projects Activity (IARPA) via Department
of Interior/Interior Business Center (DOI/IBC) contract number
D17PC00345. The U.S. Government is authorized to reproduce and
distribute reprints for Governmental purposes not withstanding any
copyright annotation theron. Disclaimer: The views and conclusions
contained herein are those of the authors and should not be
interpreted as necessarily representing the official policies or
endorsements, either expressed or implied of IARPA, DOI/IBC or
the U.S. Government.

{
\bibliographystyle{ieee_fullname}
\bibliography{meta_faster_rcnn}
}

\clearpage
\newpage
\appendix

\section*{Appendix}
The supplementary materials are organized as follows. 
First, we provide the detailed analysis for the proposed attentive feature alignment, and show more visualization examples.
Second, we provide the detailed running speed analysis of each component in our model.
Third, we provide our full experimental results on the MSCOCO FSOD benchmarks.
Fourth, we provide the implementation details of our model.

\section{Analysis of Our Proposed Attentive Feature Alignment}
\label{feature_alignment}

One of the key contributions in this paper is the introduction of the feature alignment based metric-learning method for few-shot object detection, to address the spatial misalignment between the noisy proposals and few-shot classes. 

\begin{figure}[h]
\begin{center}
\includegraphics[scale=0.24]{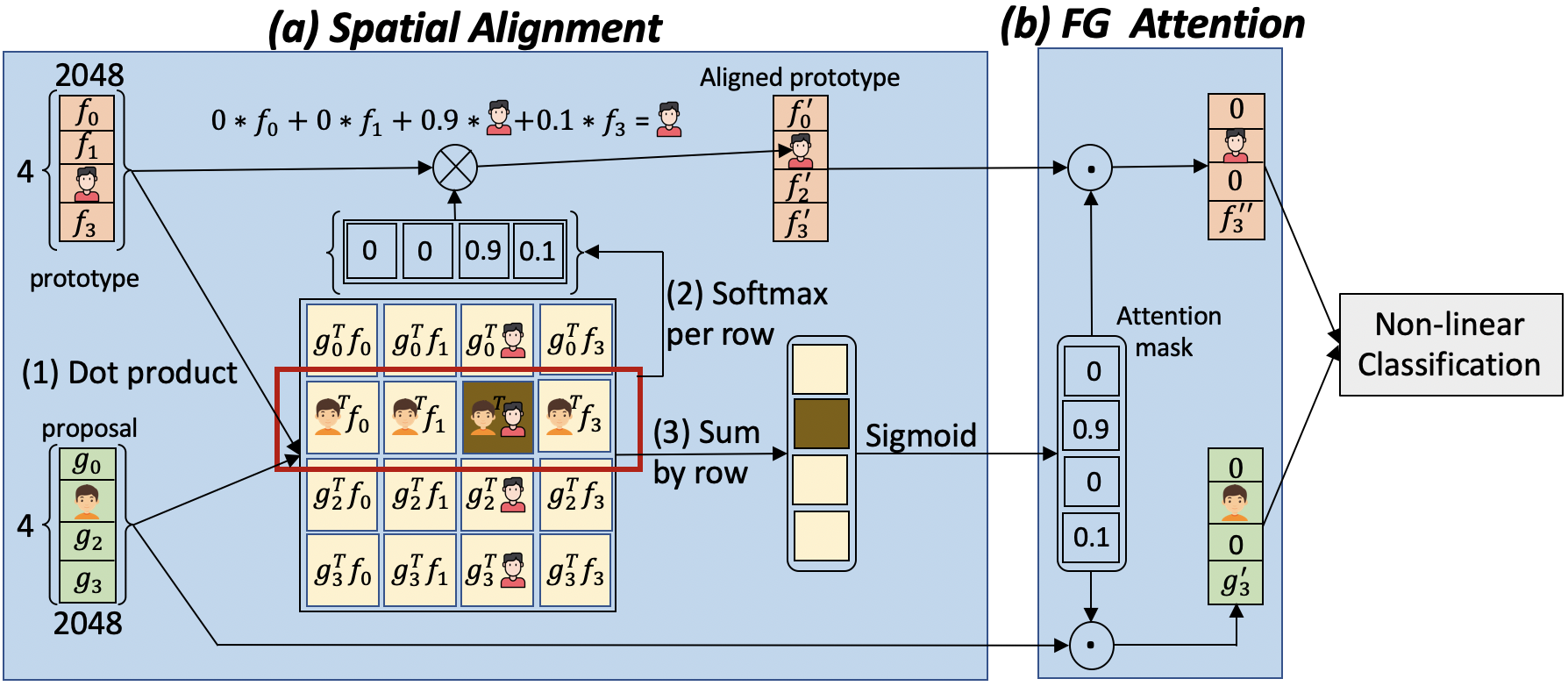}
\end{center}
\caption{The proposed attentive feature alignment. $f_0,f_1,f_3$ and $g_0,g_2,g_3$ are CNN grid features representing BG regions. $f_2$ and $g_1$ are CNN grid features with the same semantic meaning `person'. Similar semantic features are moved to the same position, and higher attention weights are assigned to the corresponding foreground regions.}
\label{Key}
\end{figure}

An example of the detailed calculation of the spatial alignment and FG attention is shown in Figure\ \ref{Key}. Formally, The dot-product in Eq.\ (2) can measure the similarity of two CNN grid features from the proposal features and class prototypes. After normalizing the affinity matrix $A$ along the row in Eq.\ (3), the prototype grid features with high similarity to the proposal grid feature in one row, also have large values (e.g. 1.0) in the normalized $A^{'}$, otherwise small values (e.g., 0.0). Then by aggregating prototype grid features with $A^{'}$ per row in Eq.\ (4), we can obtain the proposal-aligned prototype. 
Based on the affinity matrix $A$, we can calculate the probability of each CNN grid feature to be the corresponding foreground regions by aggregating $A$ along the row in Eq.\ (5). These corresponding foreground features are highlighted in Eq.\ (6) using the attention mask $M$.

Our proposed attentive feature alignment module \textbf{introduces negligible additional learning parameters}. The two newly-introduced parameters are the scaling parameters $\gamma_1$ and $\gamma_2$ in Eq.\ (7) for stable training. The remaining part in the module has no learnable parameters. This kind of design can reduce the risk of overfitting to base classes.

\begin{figure*}[t]
\begin{center}
\includegraphics[scale=0.90]{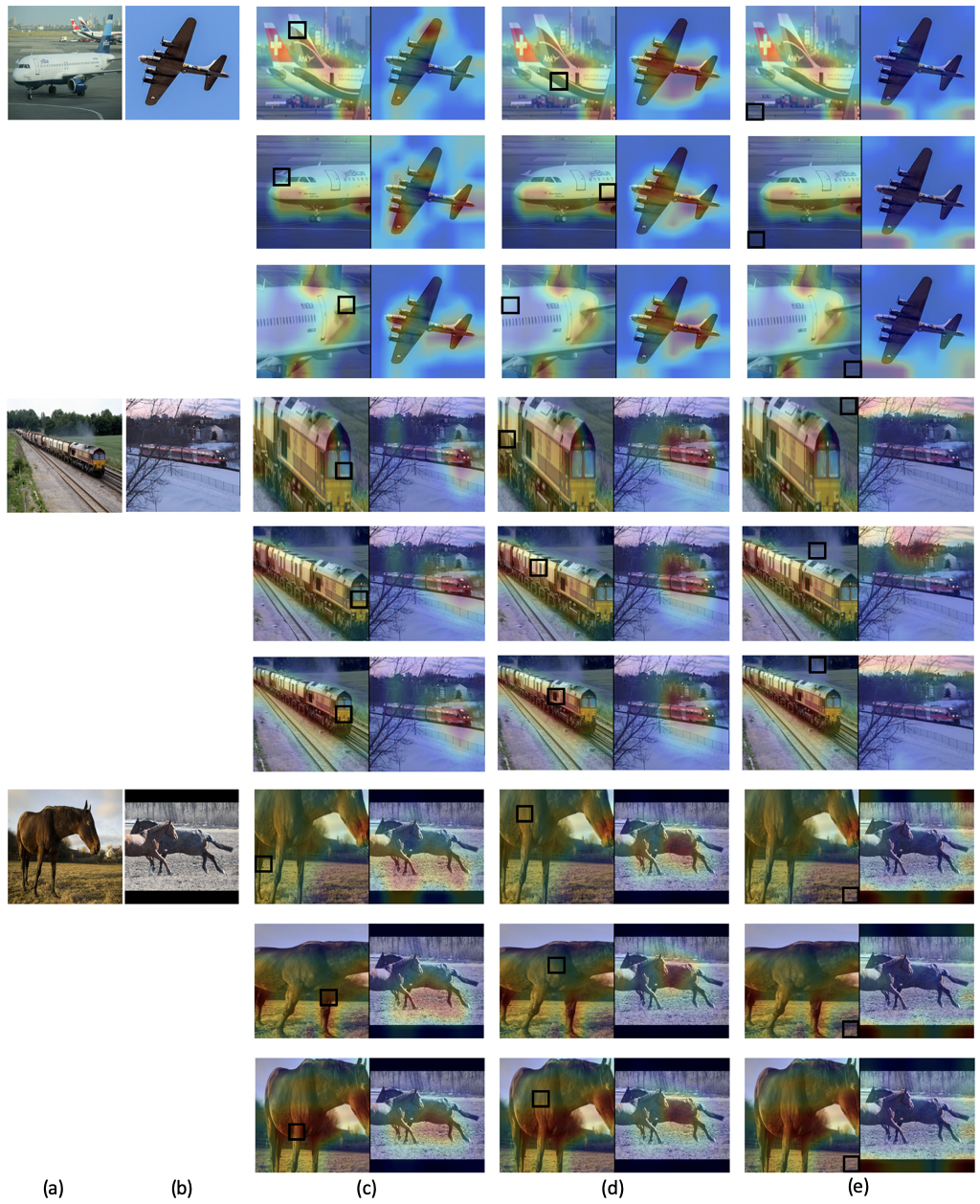}
\end{center}
\caption{(a) Query image, (b) 1-shot support image, (c)\&(d)\&(e) Left: proposal with the foreground attention mask. Right: visualization of the affinity matrix for feature alignment given the black box as query.}
\label{visualization_supp}
\end{figure*}

More visualization examples of our proposed attentive feature alignment are shown in Figure\ \ref{visualization_supp}. All these results demonstrate that the attention mask can roughly localize the foreground objects, and the affinity matrix can highlight similar semantic regions between the proposal image and 1-shot support image.

\section{Running Speed Analysis of Our Model}
\label{running_speed}

We show in Table \ref{tab_supp:running_time} the running speed (per image) of both the softmax based detector and our few-shot detector. We follow \cite{ren2015faster} to generate 1000 proposals shared by all classes in the softmax based detector, and generates 100 proposals for each class in our few-shot detector. We use different number of proposals for the two detectors because, as shown in Table \ref{tab:proposal} of the main paper, RPN could achieve similar AR (average recall) for unseen novel classes using 1000 proposals compared with our proposed Meta-RPN, which only needs 100 proposals.

From Table \ref{tab_supp:running_time} in this file and Table \ref{tab:base_class} in the main paper, we can draw the following three conclusions: \textbf{(1)} The running time of the softmax based detector is stable when detecting different number of classes. This is because most of the operations/layers are shared by different classes, and only the final linear classification layer is dependent on the number of classes. \textbf{(2)} The running time of our few-shot detector is proportional to the number of classes to detect. This is because we perform separate detection for each class, including generating class-specific proposals and performing the pairwise classification. Therefore it is inefficient to apply our few-shot detector to the base classes considering both the running speed and detection accuracy. \textbf{(3)} 
The limitation of the softmax based detector is the inflexibility to add new classes to the system because it needs to fine-tune a new classifier. On the other hand, our meta-learning method is flexible to enroll new classes without any training, which is more friendly for realistic applications. 
Considering both strengths and weaknesses of the two detectors, we propose to decouple the detection of base and novel classes into two detection heads, and use the softmax based detector for base classes and learn a class-agnostic few-shot detector for unseen novel classes. When the number of novel classes is smaller than 20, the running speed of our model is acceptable (0.6s per query image). If the number of novel classes is too much, our model could be slow. We leave the problem of speeding up our model as the future work. The model compression methods \cite{cheng2017survey} could be useful for this problem.

\begin{table*}[h]
    \centering
    \footnotesize
    \caption{Running speed (per image) of both the softmax based detector and our few-shot detector. We use ResNet-50 as our backbone network in this table. We show the running time of each components when detecting different number of classes. We use a TITAN RTX GPU for experiments.}
    \adjustbox{width=\linewidth}{
    \begin{tabular}{c|c|ccc|c}
    \toprule
    \multirow{2}{*}{Model} & \multirow{2}{*}{\#Classes to detect} & \multicolumn{3}{c|}{Running time of each component} & \multirow{2}{*}{Total running time} \\
    & & Backbone & Proposal Generation & Proposal Classification & \\\midrule
    \multirow{3}{*}{Softmax based detector} & 1 class  & 0.008s & 0.011s & 0.098s & 0.117s \\
    & 20 classes                          & 0.008s & 0.011s & 0.112s & 0.131s \\
    & 60 classes                          & 0.008s & 0.011s & 0.142s & 0.161s \\
    & 80 classes                          & 0.008s & 0.011s & 0.149s & 0.168s \\ \midrule
    \multirow{3}{*}{Our few-shot detector} & 1 class  & 0.008s & 0.011s & 0.022s & 0.041s \\
    & 20 classes                          & 0.008s & 0.169s (0.008s/class) & 0.409s (0.020s/class) & 0.586s (0.029s/class) \\ 
    & 60 classes                          & 0.008s & 0.492s (0.008s/class) & 1.292s (0.021s/class) & 1.792s (0.030s/class) \\ 
    & 80 classes                          & 0.008s & 0.643s (0.008s/class) & 1.673s (0.021s/class) & 2.324s (0.029s/class) \\ \bottomrule 
    \end{tabular}}
\label{tab_supp:running_time}
\end{table*}

\section{Full Experimental Results on the MSCOCO FSOD Benchmarks}
\label{Experimental_results}

We show in Table \ref{tab_supp:main_coco} the full evaluation results (with 1/2/3/5/10/30-shots) of our proposed Meta Faster R-CNN on the MSCOCO FSOD benchmarks, including both meta-learning and fine-tuning results. We compare our method with a strong baseline model in \cite{fan2020few}.

From both Table \ref{tab_supp:main_coco} and Table \ref{tab:main_voc} in the main file, we can draw the following three conclusions: \textbf{(1)} Both our meta-learning based model and fine-tuning model are better than the strong baseline model \cite{fan2020few}, which demonstrate the effectiveness of our model. \textbf{(2)} Our meta-learning model can even outperform the fine-tuning model of \cite{fan2020few} in low-shot settings, for example, 1-shot in the PASCAL VOC, and 1/2/3-shot in the MSCOCO. This demonstrates the strong generalization ability of our meta-learning based model. 
\textbf{(3)} Our meta-learning based model shows superior performance compared with our fine-tuning model in extremely few-shot setting, for example, 1-shot in the PASCAL VOC, and 1/2/3-shot in the MSCOCO. Fine-tuning does not help too much in low-shot settings because it is prone to over-fitting with very few samples, but could help with large shot settings if we have much more training samples (e.g., 10/30-shot). 

Therefore, our meta-learning based model is more suitable for realistic applications considering its efficiency for adaptation to novel classes without any training and its effectiveness for extremely few-shot scenarios.

\begin{table*}[ht]
\centering
\footnotesize
\addtolength{\tabcolsep}{-3pt}
\caption{Few-shot object detection performance on the novel classes of the MSCOCO dataset. We reimplement Fan et al. \cite{fan2020few} using the same few-shot training samples as \cite{kang2019few,wang2020few}. \vspace{1mm}} 
\adjustbox{width=\linewidth}{
\begin{tabular}{l|ccc|ccc|ccc|ccc|ccc|ccc}
\toprule
\multirow{2}{*}{Method}&\multicolumn{3}{c|}{1-shot} & \multicolumn{3}{c|}{2-shot} & \multicolumn{3}{c|}{3-shot} &\multicolumn{3}{c|}{5-shot} & \multicolumn{3}{c|}{10-shot} & \multicolumn{3}{c}{30-shot} \\
& AP & AP50 & AP75 & AP & AP50 & AP75 & AP & AP50 & AP75 & AP & AP50 & AP75 & AP & AP50 & AP75 & AP & AP50 & AP75 \\ \midrule
\multicolumn{19}{c}{\textbf{Meta-training the model on base classes, and meta-testing on novel classes}} \\ \midrule
Fan et al. \cite{fan2020few} & 4.0 & 8.5 & 3.5  & 5.4 & 11.6 & 4.6    & 5.9 & 12.5 & 5.0   & 6.9 & 14.3 & 6.0    & 7.6 & 15.4 & 6.8  & 8.9 & 17.8 & 8.0 \\ 
Meta Faster R-CNN &  \textbf{5.0} & \textbf{10.2} & \textbf{4.6}    & \textbf{7.0} & \textbf{13.5} & \textbf{6.4}    & \textbf{8.4} & \textbf{16.5} & \textbf{7.4}    & \textbf{9.3} & \textbf{18.1} & \textbf{8.3}   & \textbf{9.7} & \textbf{18.5} & \textbf{9.0} & \textbf{11.3} & \textbf{21.2} & \textbf{10.6} \\
\midrule

\multicolumn{18}{c}{\textbf{Fine-tuning the model on novel classes, and testing on novel classes}} \\ \midrule
Fan et al. \cite{fan2020few} & 4.2 & 9.1 & 3.0   & 5.6 & 14.0 & 3.9   & 6.6 & 15.9 & 4.9   & 8.0 & 18.5 & 6.3   & 9.6 & 20.7 & 7.7    & 13.5 & 28.5 & 11.7 \\
Meta Faster R-CNN & \textbf{5.1} & \textbf{10.7} & \textbf{4.3}  & \textbf{7.6} & \textbf{16.3} & \textbf{6.2}    & \textbf{9.8} & \textbf{20.2} & \textbf{8.2}   & \textbf{10.8} & \textbf{22.1} & \textbf{9.2}   & \textbf{12.7} & \textbf{25.7} & \textbf{10.8} & \textbf{16.6} & \textbf{31.8} & \textbf{15.8} \\
\bottomrule
\end{tabular}}
\label{tab_supp:main_coco}
\vspace{-2mm}
\end{table*}

\section{Implementation Details}
\label{Implementation_Details}

We use Faster R-CNN \cite{ren2015faster} with ResNet-50/101 \cite{he2016deep} as our base model. We use the output after $res4$ block as image features. Then RPN and Meta-RPN are used to generate proposals for base classes and novel classes respectively. After that, RoIAlign \cite{he2017mask} and the $res5$ block are used to extract features for proposals and few-shot class examples. Finally, an R-CNN classifier \cite{Fast_R-CNN} is used to perform multi-class softmax-based classification and bbox regression over the base classes, and we use the Meta-classifier to perform few-shot prototype matching and bbox regression over the novel classes. 

Next we show the details of our 3-step training framework which are omitted due to the limited space.

\textbf{Meta-learning with base classes.} In this step, we learn to perform few-shot object detection on base classes using our proposed Meta-RPN and Meta-Classifier and learn the two modules with episodic training. 
The loss function is defined as,
\begin{equation}
\label{joint_loss}
\mathcal{L}_1 = \mathcal{L}_{meta\_rpn} + \mathcal{L}_{meta\_classifier}
\end{equation}
where $\mathcal{L}_{meta\_rpn}$ and $\mathcal{L}_{meta\_classifier}$ both consist of a binary cross-entropy loss and a bbox regression loss,
\begin{align}
\mathcal{L}_{meta\_rpn} = \mathcal{L}^{B}_{cls} + \mathcal{L}_{loc}\\
\mathcal{L}_{meta\_classifier} = \mathcal{L}^{B}_{cls} + \mathcal{L}_{loc}
\end{align}
where $\mathcal{L}^{B}_{cls}$ denotes the binary cross-entropy loss over two classes, ``matched'' and ``unmatched''. The $\mathcal{L}_{loc}$ denotes the bbox regression loss using smooth $L_1$ loss defined in \cite{Fast_R-CNN}.

For model training on the MSCOCO dataset, we use the SGD optimizer with an initial learning rate of 0.002, momentum of 0.9, weight decay of 0.0001, and a batch size of 8.
To ease the learning of our model, especially the Meta-Classifier, we train our model gradually in the following three stages. First, we initialize our model with the multi-relation detector in \cite{fan2020few}. The learning rate is 0.002 in the first 30,000 iterations, and divided by 10 in the next 10,000 iterations. Second, we include the whole feature fusion network in the Meta-Classifier for training. The learning rate is 0.001 in the first 15,000 iterations, and divided by 10 in the next 5,000 iterations. Third, we include the proposed attentive feature alignment in the Meta-Classifier for training. The learning rate is 0.001 in the first 15,000 iterations, and divided by 10 in the next 5,000 iterations.

Similarly, for model training on the PASCAL VOC dataset, we use the same hyper-parameters as on the MSCOCO dataset except using fewer training iterations. The training iterations are reduced to half in all three stages. 

\textbf{Learning the separate detection head for base classes.} In this step, we learn the object detection model (including the RPN and R-CNN classifier) over base classes. We use the FPN architecture \cite{lin2017feature} to extract multi-scale features for the base-classes detection head. The loss function is defined as,
\begin{equation}
\mathcal{L}_2 = \mathcal{L}_{rpn} + \mathcal{L}_{rcnn}
\end{equation}
where $\mathcal{L}_{rpn}$ and $\mathcal{L}_{rcnn}$ both consist of a classification loss and a bbox regression loss as follows,
\begin{align}
\mathcal{L}_{rpn} = \mathcal{L}^{B}_{cls} + \mathcal{L}_{loc}\\
\mathcal{L}_{rcnn} = \mathcal{L}^{M}_{cls} + \mathcal{L}_{loc}
\end{align}
where $\mathcal{L}^{B}_{cls}$ denotes the binary cross-entropy loss over a ``foreground'' class (the union of all base classes) and a ``background'' class, and $\mathcal{L}^{M}_{cls}$ denotes the multi-class cross-entropy loss over all base classes plus a ``background'' class.

For model training on the MSCOCO dataset, we use the SGD optimizer with an initial learning rate of 0.02, momentum of 0.9, weight decay of 0.0001, and a batch size of 16. The learning rate is divided by 10 after 85,000 and 100,000 iterations. The total number of training iterations is 110,000. In this step, we fix the parameters of the backbone feature extractor so that the few-shot detection model learned in the first step does not change during this step.

For model training on the PASCAL VOC dataset, we use the same hyper-parameters as on the MSCOCO dataset except using fewer training iterations. The initial learning rate is 0.02, divided by 10 after 12,000 and 16,000 iterations. The total number of training iterations is 18,000.

\textbf{Fine-tuning with both base and novel classes.} In this step, the model is fine-tuned using a balanced dataset with both base classes and novel classes. We use the same loss function in the first step for model optimization.

For model fine-tuning on both the MSCOCO and PASCAL VOC dataset, we use the SGD optimizer with an initial learning rate of 0.001, momentum of 0.9, weight decay of 0.0001, and a batch size of 8. The learning rate is divided by 10 after 2,000 iterations, and the total number of training iterations is 3,000. The backbone feature extractor and base-classes detection head is fixed during fine-tuning.

\end{document}